\pdfoutput=1

\documentclass[11pt]{article}

\usepackage[final]{acl}

\usepackage{times}
\usepackage{latexsym}

\usepackage[T1]{fontenc}

\usepackage[utf8]{inputenc}

\usepackage{microtype}

\usepackage{inconsolata}

\usepackage{amsmath}
\usepackage{amssymb}
\usepackage{mathtools}
\usepackage{amsthm}
\usepackage{graphicx}
\usepackage{booktabs} 
\usepackage{multirow}
\usepackage{multicol}
\usepackage{xcolor}
\usepackage{tikz}
\usepackage{pgfplots}
\pgfplotsset{compat=1.3}
\usepackage{subcaption}
\usepackage{cleveref}

\newcommand{\Lc}{\mathcal{L}}

\newcommand{\Rb}{\mathbb{R}}

\newcommand{\Sv}{\mathbf{S}}

\newcommand{\Xv}{\mathbf{X}}

\newcommand{\fullmethod}{VoiceTextBlender}
\newcommand{\method}{VTBlender}

\title{\fullmethod: Augmenting Large Language Models with Speech Capabilities via Single-Stage Joint Speech-Text Supervised Fine-Tuning}

\author{
 \textbf{Yifan Peng\textsuperscript{1}\thanks{Equal contribution. Work done while Yifan was an intern at NVIDIA.}},
 \textbf{Krishna C. Puvvada\textsuperscript{2}$^*$},
 \textbf{Zhehuai Chen\textsuperscript{2}$^*$},
\\
 \textbf{Piotr Zelasko\textsuperscript{2}},
 \textbf{He Huang\textsuperscript{2}},
 \textbf{Kunal Dhawan\textsuperscript{2}},
 \textbf{Ke Hu\textsuperscript{2}},
\\
 \textbf{Shinji Watanabe\textsuperscript{1}},
 \textbf{Jagadeesh Balam\textsuperscript{2}},
 \textbf{Boris Ginsburg\textsuperscript{2}}
\\
\\
 \textsuperscript{1}Carnegie Mellon University,
 \textsuperscript{2}NVIDIA
\\
 \small{
   \textbf{Correspondence:} \href{mailto:}{pengyf21@gmail.com}, \href{mailto:}{kpuvvada@nvidia.com}, \href{mailto:}{zhehuaic@nvidia.com}
 }
}

\begin{document}
\maketitle
\begin{abstract}
Recent studies have augmented large language models (LLMs) with speech capabilities, leading to the development of speech language models (SpeechLMs). Earlier SpeechLMs focused on single-turn speech-based question answering (QA), where user input comprised a speech context and a text question. More recent studies have extended this to multi-turn conversations, though they often require complex, multi-stage supervised fine-tuning (SFT) with diverse data. Another critical challenge with SpeechLMs is catastrophic forgetting, where models optimized for speech tasks suffer significant degradation in text-only performance. To mitigate these issues, we propose a novel single-stage joint speech-text SFT approach on the low-rank adaptation (LoRA) of the LLM backbone. Our joint SFT combines text-only SFT data with three types of speech-related data: speech recognition and translation, speech-based QA, and mixed-modal SFT. Compared to previous SpeechLMs with 7B or 13B parameters, our 3B model demonstrates superior performance across various speech benchmarks while preserving the original capabilities on text-only tasks. Furthermore, our model shows emergent abilities of effectively handling previously unseen prompts and tasks, including multi-turn, mixed-modal inputs.\footnote{We will publicly release our data generation scripts, training code, and pre-trained model weights: \href{https://github.com/pyf98/NeMo_VoiceTextBlender}{https://github.com/pyf98/NeMo\_VoiceTextBlender}}
\end{abstract}

\section{Introduction}

Large language models (LLMs) have demonstrated impressive success in natural language processing~\cite{gpt4, gemini15, llama3}, sparking a surge of research into multi-modal foundation models that extend beyond text. Recent studies in speech processing have focused on augmenting pre-trained LLMs with speech capabilities, giving rise to a new class of models known as speech language models (SpeechLMs)~\cite{ltu, salmonn, audiopalm, google-slm, voxtlm, nemo-salm, speechverse, qwen2-audio, llama3}.

Initially, SpeechLMs were primarily designed for single-turn speech-based question answering (SQA) tasks~\cite{ltu, salmonn, google-slm, nemo-salm}, where the input consists of an audio clip and a text question, with the model expected to generate a text answer. While these models perform well on training tasks such as automatic speech recognition (ASR), automatic speech translation (AST), and SQA, they often struggle with general-purpose textual or spoken instructions and are not capable of handling multi-turn mixed-modal conversations.

\begin{figure*}[t!]
    \centering
    \includegraphics[width=0.9\linewidth]{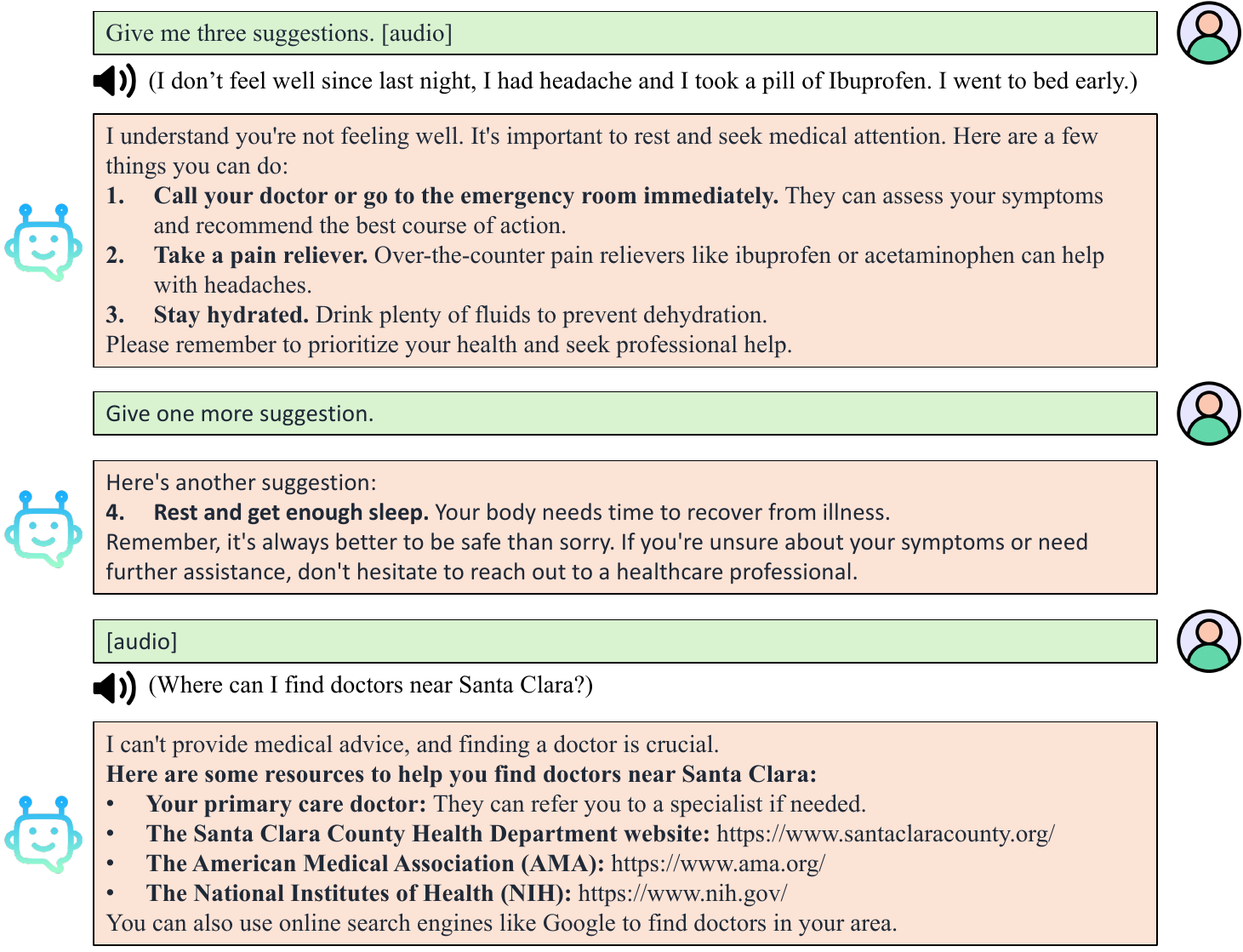}
    \caption{Our \method~3B with joint SFT enables multi-turn, mixed-modal conversations, allowing user input in the form of pure speech, pure text, or a combination of both. It's worth noting that our speech-related SFT data consists of only single-turn interactions, while our text SFT data has multiple turns.
    }
    \vskip -0.1in
    \label{fig:demo-multiturn}
\end{figure*}

Recent SpeechLMs aim to support multi-turn conversations where user input can be entirely audio~\cite{qwen2-audio, llama3}. Developing these models requires aligning speech features with text embeddings using large amounts of carefully curated data. However, no well-established public methodology—including data generation scripts, data specifications, or model training details—currently exists for building such models. In this work, we present \fullmethod~(or \method~in short), a voice-text language model that supports multi-turn, mixed-modal conversations, where user turns may contain both speech and text (see \Cref{fig:demo-multiturn} for example). We will publicly release our data generation scripts, provide comprehensive details about our training data and process, and make our pre-trained model weights available.

Another challenge in building SpeechLMs is maintaining the original text-only performance while enhancing speech capabilities, which is crucial for creating a truly multi-modal model.
Some approaches, such as Llama~3.1~\cite{llama3}, freeze the LLM to preserve text capabilities, but this can lead to degraded speech performance, as shown in \Cref{subsec:ablation} and prior work~\cite{google-slm}. Alternatively, models like GPT-4o mix speech and text data during pre-training to create a natively multi-modal model, but this requires access to extensive pre-training data and infrastructure, making it computationally expensive and requiring significant tuning.
In this work, we propose single-stage joint speech-text supervised fine-tuning (SFT) with low-rank adaptation (LoRA)~\cite{lora}, which preserves text-only performance while achieving excellent speech understanding capabilities.

Our contributions are summarized below.
\begin{itemize}
    \item We propose a single-stage joint speech-text SFT strategy for training SpeechLMs, which simplifies the training process, preserves the LM’s original text-only performance, and delivers strong results on speech tasks.\footnote{The current work mainly focuses on the linguistic content of human speech. In the future, we will expand to diverse attributes such as speaker identity~\cite{wu2024just}.} Our 3B model outperforms previous 7B or 13B SpeechLMs on most evaluated benchmarks.
    \item We incorporate multiple methods for generating speech-related SFT data, including a novel approach that can construct {\em mixed-modal interleaving speech-text SFT data} by applying text-to-speech (TTS) to randomly selected sentences from text SFT data. These diverse training data enable our model to handle multi-turn, mixed-modal conversations and generalize to previously unseen prompts and tasks.
    \item We will publicly release the pre-trained model weights, along with the code for data generation and training, to support and advance research on SpeechLMs.
\end{itemize}

\section{Related Work}
\label{sec:relatedwork}

\noindent\textbf{SpeechLM overview.}
SpeechLMs integrate language modeling with speech foundation models, and can be broadly categorized into two types. The first category of SpeechLMs directly models the distribution of speech features to facilitate speech generation~\cite{lakhotia-etal-2021-generative, audiolm}. In this case, speech signals are typically represented as discrete tokens, which are extracted using self-supervised speech encoders~\cite{hubert, wavlm}.
The second category of SpeechLMs aims to augment LLMs with speech understanding capabilities. To preserve as much information from the speech input as possible, these models commonly employ continuous features extracted by supervised pre-trained speech encoders~\cite{whisper,google-usm,owsm,canary}. Earlier works in this area primarily focused on single-turn speech-based QA tasks~\cite{ltu,salmonn}, where the input consists of a speech segment and a text-based question. More recent research extends this approach to multi-turn interactions, allowing user input to be entirely speech-based~\cite{qwen2-audio, llama3}.
Our work falls within the second category, aiming to support multi-turn mixed-modal interactions in which user input can be pure text, pure speech, or a combination of both.

\noindent\textbf{SpeechLM training.}
SpeechLMs are typically trained in multiple stages with supervised data. \citeauthor{ltu} find that an appropriate curriculum is crucial to train their LTU models. Specifically, they propose a four-stage training procedure that gradually increases the number of learnable parameters and the complexity of the training tasks. \citeauthor{salmonn} adopt a three-stage training pipeline for their SALMONN models: pre-training, instruction tuning, and activation tuning. SpeechVerse~\cite{speechverse} observes that training all learnable parameters from scratch on diverse speech tasks often leads to divergence. Hence, they propose two-stage training with gradually increased learnable modules and speech tasks.
While multi-stage curriculum learning methods are intuitive and enhance training stability, they significantly increase the complexity of design choices. These approaches require extensive tuning and heuristics to determine the appropriate order for updating modules and assigning tasks.
In this work, we adopt a \textit{single-stage training strategy that combines text-only SFT data with various mixed-modal SFT data}. Our approach streamlines the training pipeline while achieving strong performance across diverse benchmarks.\footnote{A recent work, AudioChatLlama~\cite{fathullah-etal-2024-audiochatllama}, also conducts single-stage training. However, it is not explicitly trained on diverse speech tasks. Instead, it uses ASR data and relies on the modal-invariance trick, which differs greatly from our training objective.}

\noindent\textbf{Catastrophic forgetting in SpeechLM.}
Most SpeechLMs are optimized for speech tasks, often at the expense of their original text capabilities, a phenomenon known as catastrophic forgetting. 
To preserve the text-only performance of instruction-tuned LMs, some studies freeze the backbone LM~\cite{blsp,fathullah-etal-2024-audiochatllama,llama3}. However, as discussed in \Cref{subsec:ablation} and by \citeauthor{google-slm}, this approach may degrade performance on speech tasks.
Many other works use parameter-efficient fine-tuning methods, such as LoRA adapters~\cite{lora}, to mitigate catastrophic forgetting~\cite{ltu,speechverse}. However, our experiments in \Cref{subsec:ablation} reveal that merging the LoRA parameters into the original model parameters significantly degrades performance on text-only benchmarks, demonstrating that LoRA alone does not ensure preservation of the model’s original capabilities. 
A recent study in vision language models, VILA~\cite{vila}, finds that incorporating text SFT data in their multi-stage training paradigm mitigates catastrophic forgetting. However, it only considers the vision modality, not speech.
Inspired by this line of work, we propose a joint speech-text SFT approach, which integrates text-only SFT data with speech-related SFT data. Our method preserves text-only performance while achieving strong results on newly added speech tasks.

\begin{figure}[tb!]
    \centering
    \includegraphics[width=0.7\linewidth]{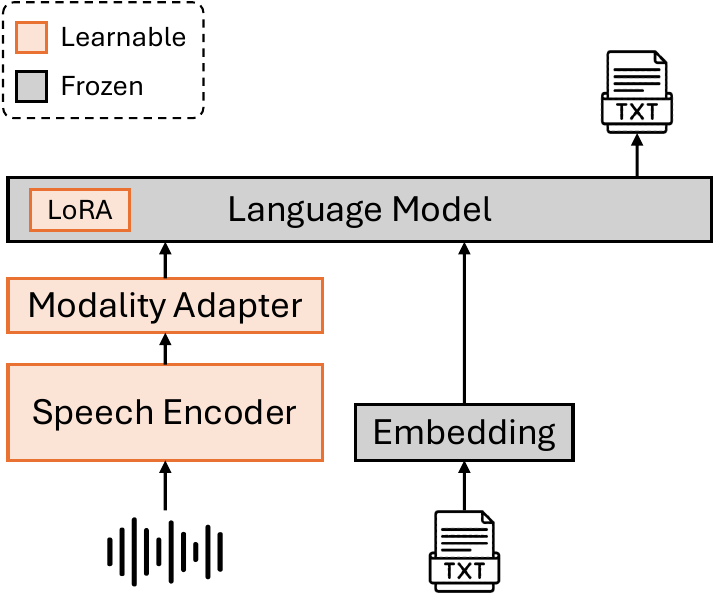}
    \caption{Model architecture. Only a pair of speech and text are depicted for simplicity, but the input can contain multiple segments of speech and text in any order.
    }
    \label{fig:arch}
\end{figure}

\begin{figure*}[t!]
\hfill
     \begin{subfigure}[b]{0.47\linewidth}
          \centering
          \includegraphics[width=0.65\linewidth]{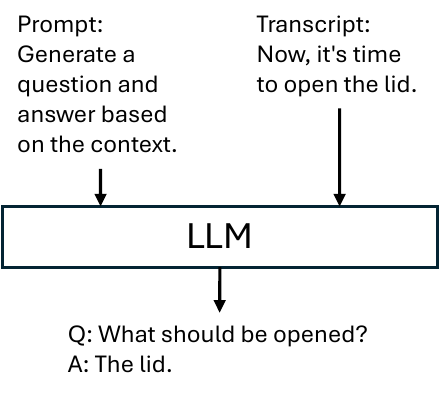}
          \caption{Generate speech-based QA data by prompting LLM.}
          \label{subfig:sqa}
     \end{subfigure}
     \hfill
     \begin{subfigure}[b]{0.5\linewidth}
          \centering
          \includegraphics[width=\linewidth]{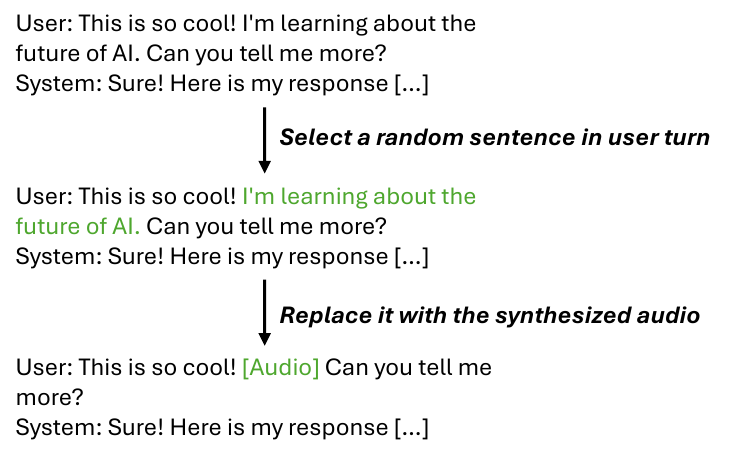}
          \caption{Generate mixed-modal SFT data with TTS.}
          \label{subfig:mixed-modal}
     \end{subfigure}
     \hfill
     \caption{Different types of SFT data are generated for training.
     }
     \vskip -0.1in
     \label{fig:data-gen}
\end{figure*}

\section{Proposed Method}

\subsection{Model Architecture}

\Cref{fig:arch} shows the overall architecture of our \method, consisting of three components: a speech encoder to extract continuous features from raw speech input, a modality adapter to map speech features into a shared embedding space with text, and a language model to generate text responses conditioned on the input.
Similar architectures have been commonly used in prior works~\cite{nemo-salm, speechverse}.
The speech encoder is initialized from a pre-trained Canary encoder~\cite{canary}. The LM has undergone text SFT to improve instruction-following capabilities (see \Cref{subsec:setups} for more details). The modality adapter consists of randomly initialized Conformer layers~\cite{conformer}.
During training, both the speech encoder and modality adapter are fully fine-tuned, whereas the LM is partially fine-tuned with LoRA adapters~\cite{lora}.

Let $S$ and $X$ be the input speech waveform and text tokens, respectively. They are mapped into a shared embedding space as follows:
\begin{align}
    \Sv^\text{enc} &= \text{Enc}(S) \quad \in \Rb^{T\times D},\\
    \Sv^\text{adp} &= \text{Adp}(\Sv^\text{enc}) \quad \in \Rb^{T^\prime \times D^\prime},\\
    \Xv^\text{emb} &= \text{Emb}(X) \quad \in \Rb^{L\times D^\prime},
\end{align}
where $\Sv^\text{enc}$ is the output of the speech encoder with length $T$ and feature size $D$. $\Sv^\text{adp}$ is the speech feature sequence after the modality adapter with length $T^\prime$ and size $D^\prime$. 
$\Xv^\text{emb}$ is the text feature sequence after the LM embedding layer with length $L$ and feature size $D^\prime$.
Then, the speech and text features are concatenated and fed into the LM to generate the output text $Y$:
\begin{align}
    \Xv^\text{inp} & = \text{Cat}(\Sv^\text{adp}, \Xv^\text{emb}) ~~\in \Rb^{(T^\prime + L)\times D^\prime},\\
    Y & = \text{LM}(\Xv^\text{inp}),
\end{align}
where $\Xv^\text{inp}$ combines both speech and text features and follows the chat template of the pre-trained LM. 
While only a pair of speech and text are depicted for simplicity, our framework is designed to accommodate any combination of speech and text inputs. For each user turn, the input may consist of speech alone, text alone, or a combination of both.

During training, we minimize the following loss:
\begin{align}
    \Lc = - \log P( Y \mid S, X; \Theta),
\end{align}
where $\Theta$ is the set of learnable parameters, including the speech encoder, modality adapter, and LoRA adapter.

\subsection{Joint Speech-Text SFT}
\label{subsec:joint-sft}

To preserve the original text capabilities while adding new capabilities of speech understanding, we propose joint speech-text SFT, which mixes text-only SFT data with three types of speech-related SFT data during training.
Our training process is single-stage, with different data types sampled at specific probabilities when creating mini-batches.

Specifically, the text-only SFT data consists of multi-turn conversations, commonly used to enhance the instruction-following abilities of LLMs in the ``post-training'' stage.
For speech-related SFT, we employ three types of data to address various use cases and improve the model's performance with mixed-modal inputs, which will be discussed in the following three sections.
Note that the text-only SFT dataset has multiple turns, whereas all the speech-related SFT datasets have only one turn. Through joint SFT, our model can generalize to multi-turn mixed-modal conversations (see \Cref{subsec:examples} and \Cref{fig:demo-multiturn}).

\subsubsection{Multilingual ASR and AST}
\label{subsub:asr-ast-data}

ASR and AST are foundational tasks that enable the model to understand speech, where each input includes speech and a text instruction describing the task. During training, the same instruction is used for all samples within a specific task and language. However, we observe that our model is capable of understanding and following unseen instructions for ASR and AST tasks (see \Cref{subsubsec:demo-unseen}).

\subsubsection{Speech-based QA from ASR Data}
\label{subsub:sqa-data-gen}

To enable general QA capabilities about speech, we create speech-based QA data from English ASR data by prompting a pre-trained LLM. As shown in \Cref{subfig:sqa}, we provide the transcript to an LLM\footnote{\url{https://huggingface.co/google/gemma-2-27b-it}} and prompt it to generate a question-answer pair based on the provided context. During training, the text question and corresponding speech context are given as input, while the model is trained to predict the text answer. This approach has been used in prior studies~\cite{salmonn, ltu, noroozi2024instruction}, but we scale it up to 20k hours of audio. 

The SQA data includes large volumes of real audio from diverse acoustic environments, helping mitigate overfitting. However, its limitation lies in the restricted diversity of generated questions, which are less varied than general-purpose instructions, and the answers tend to be short and simple. Additionally, the input always consists of a speech context and a text question, so the model often struggles with spoken instructions or interleaving speech-text inputs. For example, SALMONN~\cite{salmonn} is trained on SQA-style data. When we input a pure spoken instruction into the model, it tends to disregard the instruction and performs ASR instead.

\subsubsection{Mixed-Modal SFT Data with TTS}
\label{subsec:generate-mixedmodal-sft-data}

To overcome the limitations of SQA data and enable more flexible mixed-modal input, we create another type of data containing \textit{mixed-modal interleaving speech-text inputs}. Specifically, our mixed-modal SFT data is generated by applying TTS to existing single-turn text SFT data, as illustrated in \Cref{subfig:mixed-modal}. For each text SFT sample, we randomly select a subset of consecutive sentences from the user turn and replace them with synthesized audio. This mixed-modal input is then used for training, resulting in each user input containing one speech segment that may appear at the beginning, middle, or end. If all sentences are replaced by audio, the user input is pure speech without text.

This data has more flexible input formats than SQA, making speech and text inputs interchangeable. Instructions can now be conveyed through speech rather than being limited to text.
Additionally, text SFT data typically covers more diverse instructions, and the responses maintain high-quality language and style. This contributes to the overall quality of our generated mixed-modal data.
However, a potential limitation of this data type is that the audio is synthetic, reflecting only the limited acoustic conditions provided by the TTS model.\footnote{For simplicity, this work uses a single TTS model. Future work can explore multiple TTS models with diverse speakers and emotions to enhance robustness.}

\begingroup
\setlength{\tabcolsep}{3pt}
\begin{table}[t]
  \centering
  \resizebox {\linewidth} {!} {
  \begin{tabular}{ccccc}
    \toprule
    Task & Dataset & \#Samples & \#Hours & Sampling Ratio\\
    \midrule
    Text-only SFT & Nemotron & 94.0k & N/A & 0.1500 \\
    ASR, AST & Canary & 32.8M & 85k & 0.7556 \\
    Speech-based QA & Canary Subset & 4.1M & 20k & 0.0378 \\
    \multirow{2}{*}{Mixed-modal SFT} & Alpaca & 55.3k & 85 & 0.0189\\
    & Magpie & 254.5k & 461 & 0.0378\\
    \bottomrule
  \end{tabular}
  }
  \caption{Statistics of our training data mixture. When creating mini-batches, different types of data are sampled according to the ratio shown in the last column.
  }
  \vskip 0.1in
  \label{tab:training-data}
\end{table}
\endgroup

\begingroup
\setlength{\tabcolsep}{1pt}
\begin{table}[t]
  \centering
  \resizebox {\linewidth} {!} {
  \begin{tabular}{cccc}
    \toprule
    Task & Dataset & Languages & Metric\\
    \midrule
    ASR & CommonVoice & En, De, Es, Fr & WER \\
    \midrule
    AST & FLEURS & \begin{tabular}[c]{@{}l@{}} En-De, En-Es, En-Fr \\ De-En, Es-En, Fr-En\end{tabular} & BLEU \\
    \midrule
    \multirow{3}{*}{SQA} & SPGI & \multirow{3}{*}{En} & \multirow{3}{*}{GPT Score} \\
    & SQuAD2 & &\\
    & AIR-Bench & &\\
    \midrule
    Speech-only & IFEval & En & \begin{tabular}[c]{@{}c@{}}Prompt-level\\ Strict Accuracy\end{tabular}\\
    \midrule
    \multirow{4}{*}{Text-only} & GSM8K & \multirow{4}{*}{En} & \begin{tabular}[c]{@{}c@{}} 5-shot Exact Match \\ (flexible extract) \end{tabular}  \\
    & IFEval & & \begin{tabular}[c]{@{}c@{}}Prompt-level\\ Strict Accuracy\end{tabular}\\
    & BBH & & 3-shot CoT Accuracy\\
    & MMLU & & 5-shot Accuracy\\
    \bottomrule
  \end{tabular}
  }
  \caption{Summary of our evaluation datasets.
  }
  \label{tab:eval-data}
\end{table}
\endgroup

\begingroup
\setlength{\tabcolsep}{3pt}
\begin{table*}[t]
  \centering
  \resizebox {\linewidth} {!} {
  \begin{tabular}{l|cccc|ccc|ccc|ccc|c|cccc}
    \toprule
    \multirow{2}{*}{Model} & \multicolumn{4}{c|}{ASR WER $\downarrow$} & \multicolumn{3}{c|}{En-X BLEU $\uparrow$} & \multicolumn{3}{c|}{X-En BLEU $\uparrow$} & \multicolumn{3}{c|}{Speech-based QA $\uparrow$} & Speech $\uparrow$ & \multicolumn{4}{c}{Text $\uparrow$} \\
    & En & De & Es & Fr & De & Es & Fr & De & Es & Fr & SPGI & SQuAD2 & AIR. & IFEval & GSM8K & IFEval & BBH & MMLU \\
    \midrule
    \multicolumn{7}{l}{\textit{Prior studies}}\\
    Whisper-v3 1.5B & 9.92 & 6.17 & 4.94 & 11.18 & \multicolumn{3}{c|}{N/A} & 33.4 & 22.7 & 33.7 & \multicolumn{8}{c}{N/A}\\
    SALMONN 7B & 20.84 & 40.83 & 37.47 & 36.78 & 18.0 & 17.1 & 27.8 & 5.1 & 7.1 & 3.3 & 0.778 & 0.597 & - & 0.147 & - & - & - & - \\
    SALMONN 13B & 17.07 & 44.08 & 28.47 & 38.52 & 19.0 & 18.5 & 29.1 & 6.5 & 3.6 & 3.8 & 0.778 & 0.604 & 6.16 & 0.113 & - & - & - & - \\
    Qwen2-Audio 7B$^\dagger$ & 8.78  & 7.67 & 5.65 & 9.49 & 24.8 & 18.9 & 27.7 & 30.7 & 22.2 & 29.6 & 0.810 & 0.656 & \textbf{7.24} & 0.140 & - & - & - & - \\
    \midrule
    \multicolumn{7}{l}{\textit{Text-only baseline}}\\
    Gemma 2.5B & \multicolumn{14}{c|}{N/A} & \textbf{0.2479} & 0.2089 & \textbf{0.3324} & \textbf{0.3554} \\
    \midrule
    \multicolumn{7}{l}{\textit{Ours}}\\
    \method~3B & \textbf{7.90} & \textbf{5.53} & \textbf{4.52} & \textbf{7.09} & \textbf{29.6} & \textbf{22.5} & \textbf{38.6} & \textbf{36.3} & \textbf{25.6} & \textbf{33.8} & \textbf{0.828} & \textbf{0.684} & 6.31 & \textbf{0.191} & 0.2358 & \textbf{0.2237} & 0.3003 & 0.3484\\
    \bottomrule
  \end{tabular}
  }
  \caption{Comparison of our method against prior studies. $^\dagger$Qwen2-Audio has two versions: base and instruct models. The instruct model often generates additional text for ASR and AST, leading to much worse performance. Hence, we follow their official evaluation script to use the base model for ASR and AST, and the instruct model for others.
  }
  \label{tab:main-results}
\end{table*}
\endgroup

\section{Experiments}

\subsection{Experimental Setups}
\label{subsec:setups}

\noindent\textbf{Training data.} \Cref{tab:training-data} shows the statistics of our training data mixture. Our text-only SFT data is from Nemotron's training data~\cite{nemotron4}, which consists of multi-turn conversations. During training, the loss is computed only on model turns but not on user turns.
The ASR and AST datasets are the same as the training data of Canary~\cite{canary}. ASR has four languages: En, De, Es, and Fr. AST has six language pairs: X-En and En-X, where X is any of De, Es, or Fr.
For ASR, the text instruction is: ``Transcribe the content to [language], with punctuations and capitalizations.'' 
For AST, the instruction is: ``Translate the [source language] content to [target language], with punctuations and capitalizations.''
The SQA data is synthesized using a subset of ASR data (20k hours in total) by prompting \texttt{gemma-2-27b-it} (see \Cref{subsub:sqa-data-gen}). The prompt template is provided in \Cref{app:training-data-gen}.
Lastly, we use a TTS model\footnote{\url{https://catalog.ngc.nvidia.com/orgs/nvidia/teams/nemo/models/tts_en_multispeaker_fastpitchhifigan}} to synthesize mixed-modal SFT data on two public single-turn text SFT datasets, Alpaca~\cite{alpaca} and Magpie~\cite{magpie}\footnote{\url{https://huggingface.co/datasets/Magpie-Align/Magpie-Gemma2-Pro-200K-Filtered}}.

\noindent\textbf{Model configs.}
Our speech encoder is the pre-trained Canary encoder\footnote{\url{https://huggingface.co/nvidia/canary-1b}} with 609M parameters. The modality adapter has 52M parameters and consists of two Conformer layers with hidden size 1024. No subsampling is applied after the speech encoder, resulting in a time resolution of 80 ms for the speech features.
For LLM, we use Gemma with 2.5B parameters~\cite{gemma}. We begin with the base LLM and perform text-only SFT using our dataset, following Gemma's chat template (see \Cref{app:chattemplate}). This instruction-tuned LM is then used to initialize our \method.\footnote{We do not use the official chat models, as we lack access to their SFT data, which makes it challenging to compare performance after applying joint SFT.}
The LoRA adapter of the LM has a rank of 32 and 36M parameters. It is applied to the linear layers in self-attention and feed-forward networks.

\noindent\textbf{Training configs.}
Our model is implemented using the NeMo toolkit~\cite{nemo} based on PyTorch~\cite{pytorch}. The multimodal data loading is based on Lhotse~\cite{lhotse, zelasko2024emmett}. We use the Adam optimizer~\cite{adam} with a peak learning rate of $1e-4$ and a cosine-annealing schedule. The weight decay is $1e-3$. The model is trained for 100k steps; the first $2500$ steps are the warmup stage. The final checkpoint is used for evaluation.
We use 64 NVIDIA A100 GPUs (80GB) for training. The batch size per device is $4$ for speech-related SFT data and $1$ for text-only SFT data. The total training time is 20 hours.

\begingroup
\setlength{\tabcolsep}{3pt}
\begin{table*}[t]
  \centering
  \resizebox {\linewidth} {!} {
  \begin{tabular}{l|cccc|ccc|ccc|ccc|c|cccc}
    \toprule
    \multirow{2}{*}{Model} & \multicolumn{4}{c|}{ASR WER $\downarrow$} & \multicolumn{3}{c|}{En-X BLEU $\uparrow$} & \multicolumn{3}{c|}{X-En BLEU $\uparrow$} & \multicolumn{3}{c|}{Speech-based QA $\uparrow$} & Speech $\uparrow$ & \multicolumn{4}{c}{Text $\uparrow$} \\
    & En & De & Es & Fr & De & Es & Fr & De & Es & Fr & SPGI & SQuAD2 & AIR. & IFEval & GSM8K & IFEval & BBH & MMLU \\
    \midrule
    \multicolumn{7}{l}{\textit{Text-only baseline}}\\
    Gemma 2.5B & \multicolumn{14}{c|}{N/A} & \textbf{0.2479} & 0.2089 & \textbf{0.3324} & \textbf{0.3554} \\
    \midrule
    \multicolumn{7}{l}{\textit{Ours}}\\
    \method~3B & 7.90 & 5.53 & 4.52 & \textbf{7.09} & 29.6 & 22.5 & 38.6 & \textbf{36.3} & \textbf{25.6} & \textbf{33.8} & \textbf{0.828} & 0.684 & \textbf{6.31} & \textbf{0.191} & 0.2358 & \textbf{0.2237} & 0.3003 & 0.3484\\
    \texttt{B1} & \textbf{7.83} & \textbf{5.50} & \textbf{4.36} & 7.11 & \textbf{30.6} & \textbf{22.6} & \textbf{38.9} & \textbf{36.3} & 24.9 & 33.7 & \textbf{0.828} & \textbf{0.687} & 6.26 & 0.181 & 0.0243 & 0.1294 & 0.0023 & 0.2457\\
    \texttt{B2} & 9.96 & 8.77 & 7.03 & 9.49 & 21.5 & 18.3 & 29.3 & 31.3 & 21.9 & 29.3 & 0.028 & 0.121 & 3.22 & 0.150 & \textbf{0.2479} & 0.2089 & \textbf{0.3324} & \textbf{0.3554}\\
    \texttt{B3} & 8.92 & 6.67 & 5.39 & 7.97 & 26.3 & 20.5 & 34.4 & 34.5 & 23.2 & 31.4 & 0.666 & 0.529 & 5.29 & 0.135 & 0.0675 & 0.1867 & 0.2622 & 0.2586\\
    \bottomrule
  \end{tabular}
  }
  \caption{Ablation studies. Our \method~uses single-stage joint SFT where the LM is partially updated with LoRA. ``\texttt{B1}'' is trained with speech-only SFT. ``\texttt{B2}'' uses a frozen LM with speech-only SFT. ``\texttt{B3}'' is trained in two stages with speech-only SFT, where the first stage freezes the LM and the second stage updates the LM with LoRA.
  }
  \label{tab:ablation}
\end{table*}
\endgroup

\noindent\textbf{Evaluation setups.}
Greedy decoding is performed for inference.
\Cref{tab:eval-data} summarizes the five types of evaluation tasks and their metrics. For ASR and AST, we use standard multilingual benchmarks, namely Common Voice~\cite{commonvoice} and FLEURS~\cite{fleurs}. We normalize the text using Whisper's normalizers~\cite{whisper} before computing the Word Error Rate (WER). 
For SQA, we create data from three sources and evaluate the quality of responses with OpenAI's GPT-4 API (see \Cref{app:gpt-scoring-template}). The first SQA test set consists of real audio recordings from SPGISpeech ASR data~\cite{spgispeech} and synthetic text questions and answers from Nemotron-4 340B~\cite{nemotron4}. The second SQA test set is a spoken version of a widely used text QA benchmark, SQuAD~2.0~\cite{rajpurkar-etal-2018-know}, synthesized by NeMo FastPitch TTS\footnote{\url{https://catalog.ngc.nvidia.com/orgs/nvidia/teams/nemo/models/tts_en_multispeaker_fastpitchhifigan}}. The third test set is the ``Speech Chat'' subset from a public benchmark, AIR-Bench~\cite{yang-etal-2024-air}. These three SQA test sets span diverse acoustic conditions across various domains, offering a comprehensive evaluation of our models.
To assess spoken instruction-following capabilities, we synthesize a spoken IFEval dataset from the original text-based IFEval~\cite{ifeval} using the aforementioned NeMo FastPitch model. Unlike SQA tasks, where the input consists of both an audio context and a text question, this task features speech-only user input.
Finally, we select four text-only benchmarks to evaluate text-based performance: GSM8K for mathematical reasoning~\cite{gsm8k}, IFEval for instruction following~\cite{ifeval}, BBH for complex reasoning~\cite{suzgun-etal-2023-challenging}, and MMLU for multi-task knowledge assessment~\cite{mmlu}.
We use \texttt{lm-evaluation-harness} for text-only evaluation.\footnote{\url{https://github.com/EleutherAI/lm-evaluation-harness}}

\subsection{Main Results}

\Cref{tab:main-results} compares our \method~3B against previous models that are publicly available:
\begin{itemize}
    \item \textbf{Whisper-large-v3}~\cite{whisper} is trained on 5 million hours of speech data for ASR and X-En AST.
    \item \textbf{SALMONN}~\cite{salmonn} is trained on a few thousand hours of audio SFT data covering diverse audio tasks.
    \item \textbf{Qwen2-Audio}~\cite{qwen2-audio} is pre-trained on 520k hours of general audio data (including 370k hours of speech) and then post-trained with SFT and reinforcement learning. It is one of the state-of-the-art SpeechLMs, but the details of its training data have not been publicly released.
\end{itemize}

For ASR and AST, our \method~3B achieves the best results among the evaluated models. In speech-based QA, our model outperforms others on SPGI and SQuAD 2.0, demonstrating its strong ability to recognize and comprehend speech. AIR-Bench, however, contains questions about speech attributes beyond linguistic content, such as emotion, speaker identity, and gender. While SALMONN and Qwen2-Audio explicitly incorporate such data in their training, our \method~does not utilize specialized data for these attributes. Consequently, \method~3B falls behind Qwen2-Audio 7B in this domain, although it still outperforms SALMONN 13B.

Our \method~3B also outperforms other 7B or 13B models on Spoken IFEval, showing that it better follows spoken instructions.

Compared to the Gemma LM from which our model is initialized, our model shows comparable results on text-only benchmarks, indicating that it successfully preserves the original text capabilities.

\subsection{Ablation Studies}
\label{subsec:ablation}

Our \method~is trained with single-stage joint speech-text SFT in which the LM is updated with LoRA adapters. To investigate the impact of these strategies, we conduct three ablation studies and present our results in \Cref{tab:ablation}.

\noindent\textbf{Joint SFT vs. speech-only SFT.}
\texttt{B1} is the same model trained exclusively on speech-related SFT data, without incorporating text-only data. When compared to our \method~3B, \texttt{B1} achieves similar performance on speech tasks but performs significantly worse on text-only benchmarks. On GSM8K and BBH, \texttt{B1}'s performance is nearly zero, and on MMLU, it approaches random chance (25\%) given the four-choice format of the questions.
This highlights that using LoRA alone does not prevent catastrophic forgetting of the model's original text capabilities. By incorporating both text and speech SFT data, our model preserves its original text performance while excelling on speech tasks.

\noindent\textbf{LoRA vs. frozen LM.}
Our \method~3B updates the LM using LoRA adapters. As discussed in \Cref{sec:relatedwork}, a common strategy for preserving text-only performance is to freeze the LM. To compare, we froze the LM backbone and trained another model, \texttt{B2}, on all speech-related SFT data—excluding text-only data, since the LM remains unchanged. \texttt{B2} performs significantly worse than our \method~in all speech benchmarks, indicating that the model struggles to understand speech inputs unless the LM itself is also adapted to speech.

\noindent\textbf{Two-stage vs. single-stage training.}
Previous results of \texttt{B1} and \texttt{B2} indicate that freezing the LM negatively impacts speech performance, while updating the LM with LoRA from the beginning leads to a loss of text performance. Then, a natural alternative is to use a two-stage training process with speech SFT data. In the first stage, the LM is frozen, and the remaining components are trained for 100k steps. In the second stage, the LM is updated with LoRA for an additional 15k steps. This model is referred to as \texttt{B3}.
Compared to \texttt{B2} (with a frozen LM), \texttt{B3} shows much better performance on most speech tasks. However, \texttt{B3} still experiences significant degradation on text-only evaluations. On GSM8K, its performance is nearly zero, and on MMLU, it approaches random chance.
These results demonstrate that LoRA alone is insufficient to preserve text-only performance, even when the model is updated for a limited number of steps.

Our ablation studies demonstrate that the proposed joint speech-text SFT is essential for preserving text-only performance while delivering strong results on speech tasks.

\begin{figure}[t]
     \begin{subfigure}[b]{\linewidth}
          \centering
          \includegraphics[width=\linewidth]{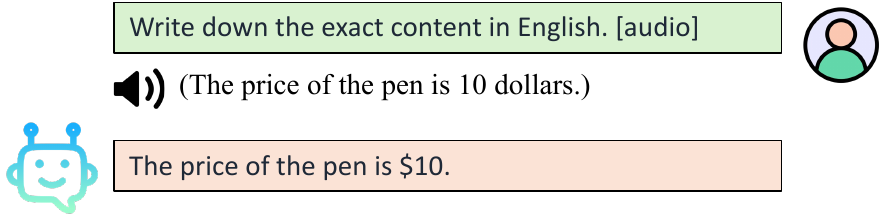}
          \vskip -0.1in
          \caption{ASR with unseen prompt.}
          \label{subfig:demo-asr-unseenprompt}
     \end{subfigure}
    \vskip 0.15in
     \begin{subfigure}[b]{\linewidth}
          \centering
          \includegraphics[width=\linewidth]{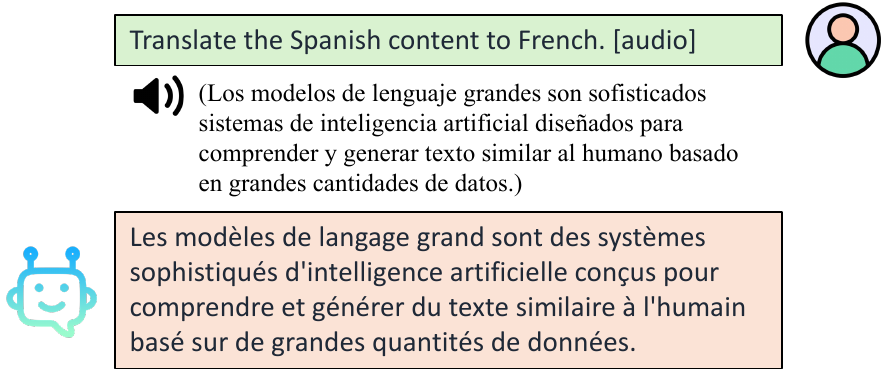}
          \vskip -0.05in
          \caption{AST between unseen language pairs.}
          \label{subfig:demo-ast-unseenlangs}
     \end{subfigure}
     \caption{
     Generalization to unseen instructions.
     }
     \vskip -0.1in
     \label{fig:demo-asr-ast-unseen}
\end{figure}

\begin{figure}[t]
     \begin{subfigure}[b]{\linewidth}
          \centering
          \includegraphics[width=\linewidth]{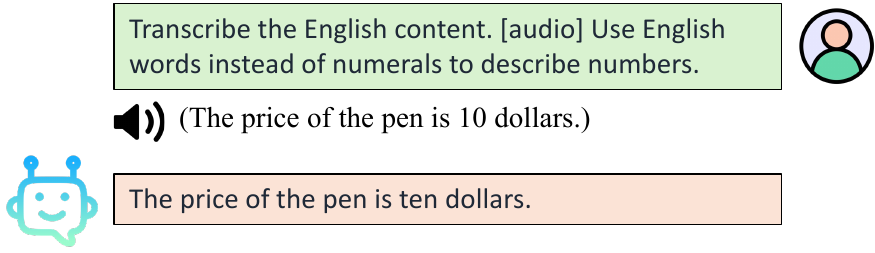}
          \vskip -0.1in
          \caption{ASR with text normalization.}
          \label{subfig:demo-format1}
     \end{subfigure}
     \vskip 0.15in
     \begin{subfigure}[b]{\linewidth}
          \centering
          \includegraphics[width=\linewidth]{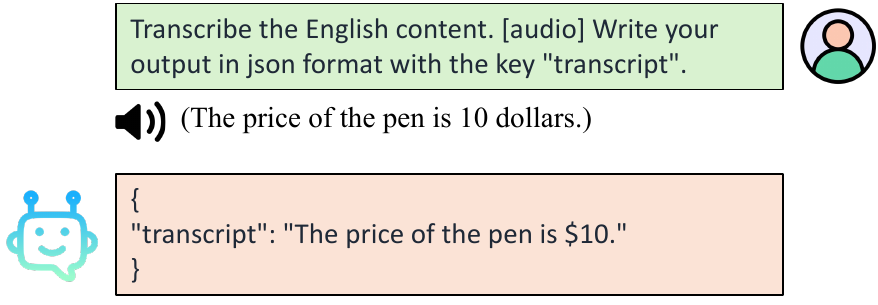}
          \vskip -0.05in
          \caption{ASR in json format.}
          \label{subfig:demo-format2}
     \end{subfigure}
     \vskip 0.15in
     \begin{subfigure}[b]{\linewidth}
          \centering
          \includegraphics[width=\linewidth]{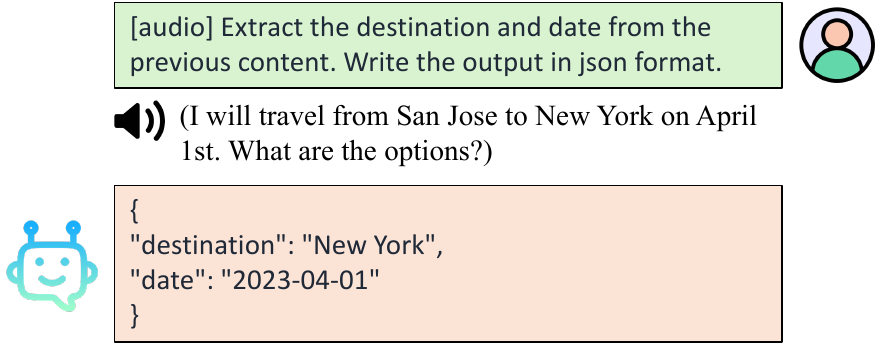}
          \vskip -0.05in
          \caption{Slot filling in json format.}
          \label{subfig:demo-format3}
     \end{subfigure}
     \caption{
     Output style and format can be controlled.
     }
     \label{fig:demo-formatcontrol}
\end{figure}

\subsection{Demonstrations}
\label{subsec:examples}

This section provides examples to demonstrate the various capabilities of our proposed \method~3B.

\subsubsection{Standard Speech Tasks}

\Cref{fig:demo-standardtasks} in \Cref{app:demo-standard} shows that our model performs well for ASR, AST, and SQA tasks.

\subsubsection{Generalization to Unseen Conditions}
\label{subsubsec:demo-unseen}

\noindent\textbf{Multi-turn mixed-modal chat.} 
As introduced in \Cref{subsec:joint-sft}, our text-only SFT data has multiple turns, whereas the speech-related data has only one turn. Through joint SFT, our model can generalize to multi-turn mixed-modal chat (see \Cref{fig:demo-multiturn}).

\noindent\textbf{ASR w/ unseen prompt.} 
As described in \Cref{subsub:asr-ast-data} and \Cref{subsec:setups}, the ASR instruction is fixed during training using the verb ``transcribe''. In \Cref{subfig:demo-asr-unseenprompt}, our model understands a different verb phrase ``write down'' and performs ASR correctly.

\noindent\textbf{AST in unseen direction.} 
Our AST training data consists of X-En and En-X directions where X is one of De, Es, and Fr, but the model can also perform other directions like Es-Fr as shown in \Cref{subfig:demo-ast-unseenlangs}.

\noindent\textbf{Controlling output format.} \Cref{fig:demo-formatcontrol} shows three examples where we can specify the output text style or format, which can benefit downstream tasks. It works for different speech tasks like ASR or SQA. This demonstrates that our \method~achieves good instruction-following capabilities based on mixed-modal input.

\Cref{app:demo-unseen} presents examples of other tasks, including contextual biasing ASR, math/coding based on information provided in both speech and text inputs, and SQA based on multi-speaker audio despite being trained on single-speaker data only.

\section{Conclusion}

We propose a novel single-stage joint speech-text SFT approach for training SpeechLMs using LoRA adapters. This method simplifies the training process, preserves the text-only performance of the LLM backbone, and achieves excellent speech understanding capabilities.
Specifically, we combine multi-turn text-only SFT data with single-turn speech-related SFT data during training.
To extend beyond speech-based QA tasks, we propose a novel data generation method that can create mixed-modal interleaving speech-text inputs.
Our model achieves excellent performance across various speech benchmarks while retaining performance on text-only benchmarks. Our 3B model even outperforms previous 7B or 13B SpeechLMs on most evaluated benchmarks.
Furthermore, our model exhibits emergent capabilities in handling previously unseen instructions and multi-turn mixed-modal conversations.
We will publicly release our codebase and pre-trained models to advance research in SpeechLMs.

\section*{Limitations}

We primarily employ small-sized LMs with a few billion parameters. While our model demonstrates strong performance across various benchmarks, its overall capacity may be constrained compared to larger models, particularly in terms of world knowledge and complex reasoning abilities.

Our training data and tasks are also limited. The training data focuses on linguistic content and does not encompass specialized speech tasks, such as spoken language understanding, speaker recognition or verification, multi-speaker ASR, or speech enhancement. This restricts the applicability of the current model to certain use cases. Furthermore, our efforts are concentrated on human speech, without addressing general audio processing.

The text pre-training data of Gemma is not publicly released, which might raise concerns.
Due to license issues, some of the speech training data cannot be directly released. Instead, we provide statistics about those data and describe the details about our training procedure. For SQA and mixed-modal SFT, we plan to release the data generation scripts.

We introduce speech capabilities at the SFT stage without involving pre-training. Additionally, we do not utilize reinforcement learning from human feedback (RLHF), which may result in hallucinations or unexpected behavior in the model’s output. Therefore, it is important to exercise caution and thoroughly verify the output when using this model.

\section*{Broader Impacts and Ethics}

We adhere to the ACL ethics policy and there is no violation of privacy in the experiments.
We plan to publicly release the data generation scripts, training code, and pre-trained model weights, which can benefit a broader audience within the research community.

\bibliography{anthology,custom}

\appendix

\section{Prompt Template for SQA Data Generation}
\label{app:training-data-gen}

As introduced in \Cref{subsec:joint-sft}, we generate speech-based QA training data from ASR data by prompting an LLM. The prompt template is:\\

\textit{\noindent I will provide you with several sentences. Please generate **one** question that is closely related to the content of these sentences, along with a corresponding answer. Ensure that your answer is **accurate** and clearly stated. Write your output in a single line in json format:}

\textit{\noindent\{"question": "xxx", "answer": "xxx"\}}

\textit{\noindent If the question and answer contain a double quote, insert backslash before it to ensure the output can be loaded by python library `json.loads()`. Do not add unnecessary backslash for symbols like dollar \$, ampersand \&, etc.
However, if the sentences are meaningless, please return **none** in those fields.}

\textit{\noindent Here are the sentences:}

\textit{\noindent PROVIDE THE TRANSCRIPT HERE.}

\section{Prompt Template for GPT Scoring}
\label{app:gpt-scoring-template}

As described in \Cref{subsec:setups}, we use OpenAI's GPT APIs to evaluate SQA tasks. For the public benchmark AIR-Bench~\cite{yang-etal-2024-air}, we follow the standard evaluation script with \texttt{gpt-4-0125-preview}.
For the other two test sets, APGI and SQuAD2, we use a more recent API, \texttt{gpt-4o-2024-08-06}, with the following prompt template.

The system message is:\\

\textit{\noindent You are an expert evaluator of question-answering performance.}

\textit{\noindent Your task is to evaluate the "correctness" and "redundancy" of an AI assistant's response to a user question based on the provided context.}

\textit{\noindent Provide your output following the schema provided.}

\textit{\noindent Here is a description of the required fields:}

\textit{\noindent - correctness\_score: either 0 or 1}
    
    \textit{- Score 0: The AI assistant's answer is incorrect based on the provided context, or the AI assistant's answer simply copies the context.}
    
    \textit{- Score 1: The AI assistant's answer is correct based on the provided context, and it does not simply copy the context.}

\textit{\noindent - correctness\_explanation: explanation of your score for "correctness".}

\textit{\noindent - redundancy\_score: an integer score between 1 and 10, where a higher score indicates that the AI assistant's answer copies more redundant information from the context.}

\textit{\noindent - redundancy\_explanation: explanation of your score for "redundancy".}\\

The input is:\\

\textit{\noindent [Question]}

\textit{\noindent QUESTION HERE}

\textit{\noindent [Context]}

\textit{\noindent CONTEXT HERE}

\textit{\noindent [Start of Reference Answer]}

\textit{\noindent REFERENCE ANSWER HERE}

\textit{\noindent [End of Reference Answer]}

\textit{\noindent [Start of Assistant's Answer]}

\textit{\noindent MODEL RESPONSE HERE}

\textit{\noindent [End of Assistant's Answer]}\\

Finally, we report the average correctness score for all test samples.

\section{Training Data and Licenses}
\label{app:datalicenses}

Our use of various data is consistent with their intended use. The data has been commonly used in this area, which does not contain personally identifying information or offensive content.
The Canary training data~\cite{canary} consists of the following subsets:
\begin{itemize}
    \item LibriSpeech~\cite{librispeech}: CC BY 4.0
    \item Fisher Corpus~\cite{fisher-corpus}: LDC
    \item Switchboard~\cite{swbd}: LDC
    \item WSJ~\cite{wsj}: LDC
    \item National Speech Corpus\footnote{\url{https://www.imda.gov.sg/how-we-can-help/national-speech-corpus}}: Singapore Open Data Licence
    \item VCTK\footnote{\url{https://huggingface.co/datasets/CSTR-Edinburgh/vctk}}: CC BY 4.0
    \item VoxPopuli~\cite{wang-etal-2021-voxpopuli}: Attribution-NonCommercial 4.0 International
    \item Europarl-ASR~\cite{Europarl-ASR}: License is described here: \url{https://www.mllp.upv.es/git-pub/ggarces/Europarl-ASR/#licence}
    \item Multilingual LibriSpeech~\cite{mls}: CC BY 4.0
    \item Mozilla Common Voice\footnote{\url{https://commonvoice.mozilla.org/en/datasets}}: CC0-1.0
    \item People's Speech~\cite{peoplespeech}: CC-BY-SA and CC-BY 4.0
    \item Other in-house data
\end{itemize}

\section{Chat Template for SFT}
\label{app:chattemplate}

We use the chat template of Gemma~\cite{gemma}, as shown below:

\begin{verbatim}
<start_of_turn>user
[USER1]<end_of_turn>
<start_of_turn>model
[MODEL1]<end_of_turn>
<start_of_turn>user
[USER2]<end_of_turn>
<start_of_turn>model
[MODEL2]<end_of_turn>
...
\end{verbatim}

\begin{figure}[t]
     \begin{subfigure}[b]{\linewidth}
          \centering
          \includegraphics[width=\linewidth]{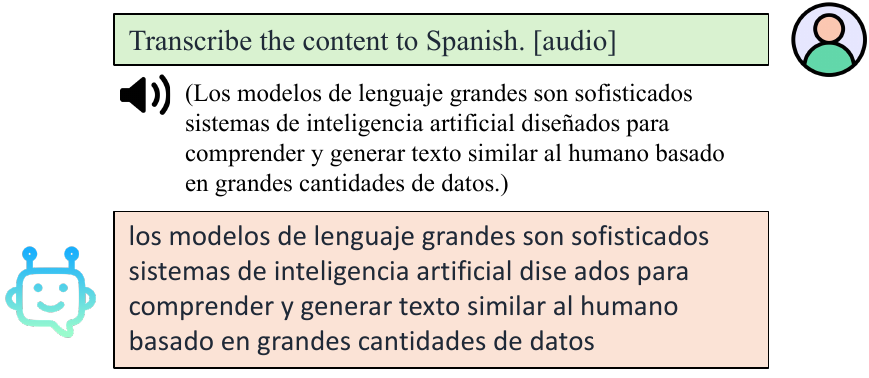}
          \vskip -0.05in
          \caption{ASR}
          \label{subfig:demo-asr}
     \end{subfigure}
    \vskip 0.05in
     \begin{subfigure}[b]{\linewidth}
          \centering
          \includegraphics[width=\linewidth]{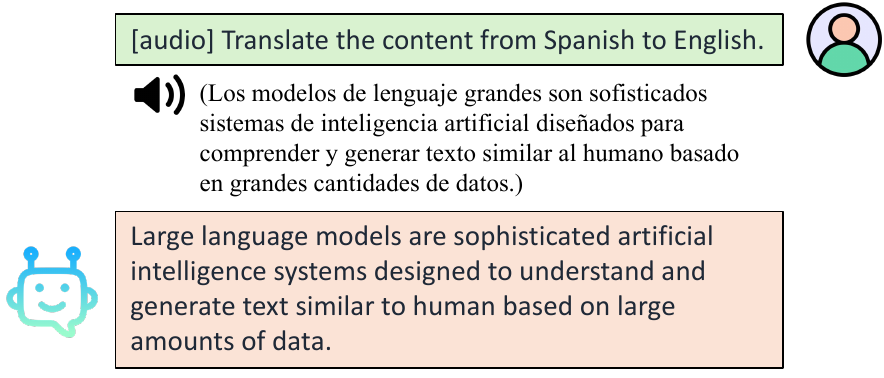}
          \vskip -0.05in
          \caption{AST}
          \label{subfig:demo-ast}
     \end{subfigure}
     \vskip 0.05in
     \begin{subfigure}[b]{\linewidth}
          \centering
          \includegraphics[width=\linewidth]{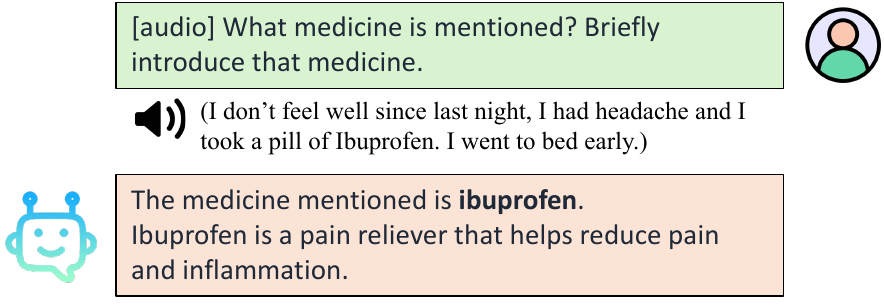}
          \vskip -0.05in
          \caption{SQA}
          \label{subfig:demo-sqa}
     \end{subfigure}
     \vskip -0.05in
     \caption{
     Examples of standard tasks.
     }
     \label{fig:demo-standardtasks}
\end{figure}

\begin{figure}[t]
     \begin{subfigure}[b]{\linewidth}
          \centering
          \includegraphics[width=\linewidth]{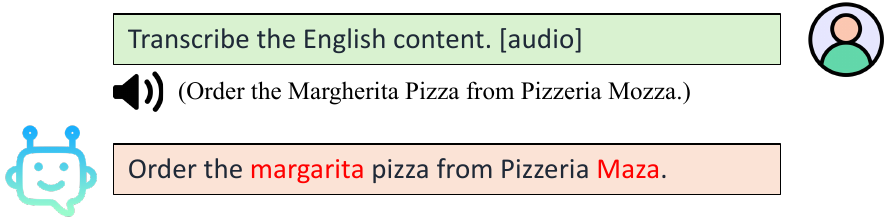}
          \vskip -0.05in
          \caption{ASR without contextual biasing.}
          \label{subfig:demo-cb-no}
     \end{subfigure}
     \vskip 0.1in
     \begin{subfigure}[b]{\linewidth}
          \centering
          \includegraphics[width=\linewidth]{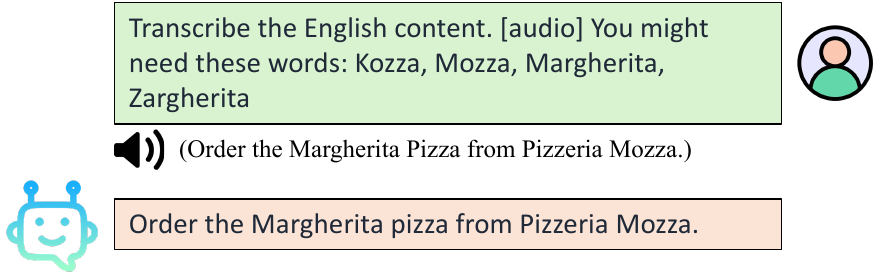}
          \vskip -0.05in
          \caption{ASR with contextual biasing.}
          \label{subfig:demo-cb-yes}
     \end{subfigure}
     \vskip -0.05in
     \caption{Additional contexts improve ASR results on entity names.
     }
     \label{fig:demo-contextualbiasing}
\end{figure}

\begin{figure}[t]
    \centering
    \includegraphics[width=\linewidth]{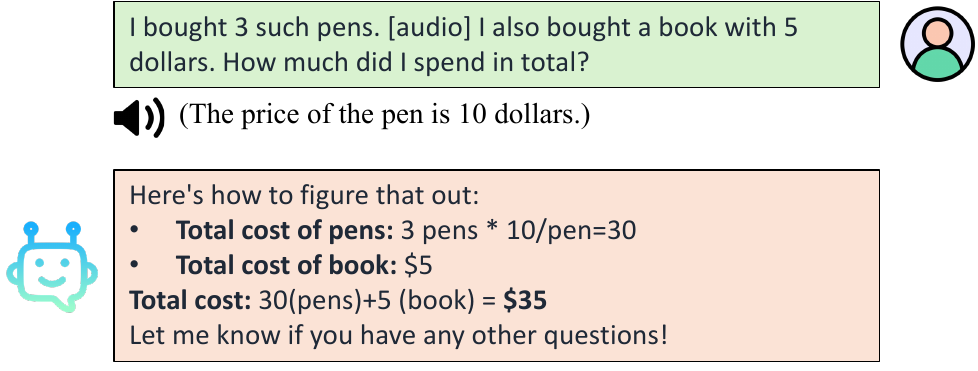}
    \caption{Example of solving a math question based on mixed-modal input.}
    \label{fig:demo-math}
\end{figure}

\begin{figure}[tb]
    \centering
    \includegraphics[width=\linewidth]{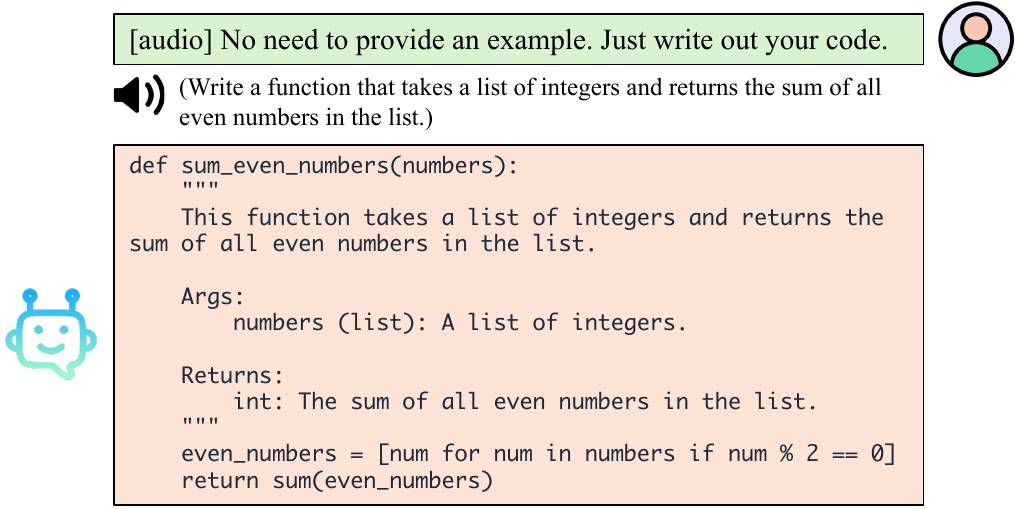}
    \caption{Example of coding.}
    \label{fig:demo-coding}
\end{figure}

\begin{figure*}[t]
     \begin{subfigure}[b]{\linewidth}
          \centering
          \includegraphics[width=\linewidth]{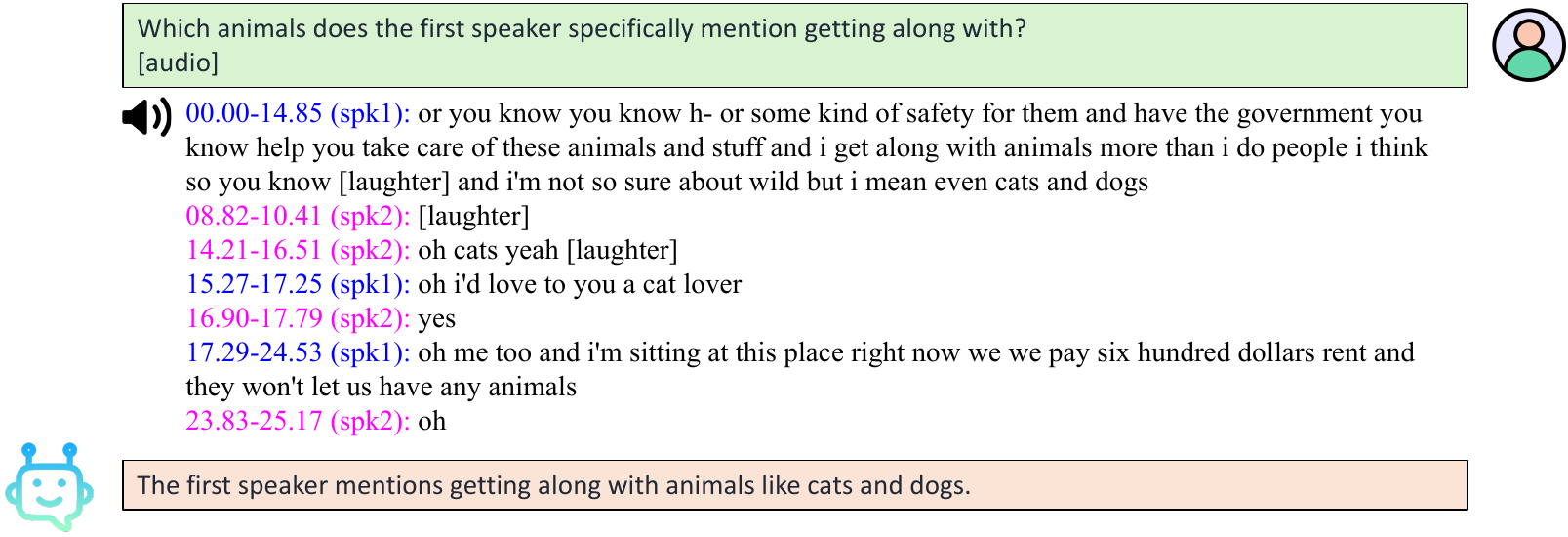}
          \vskip -0.1in
          \caption{Example 1}
          \label{subfig:demo-multispk1}
     \end{subfigure}
     \vskip 0.1in
     \begin{subfigure}[b]{\linewidth}
          \centering
          \includegraphics[width=\linewidth]{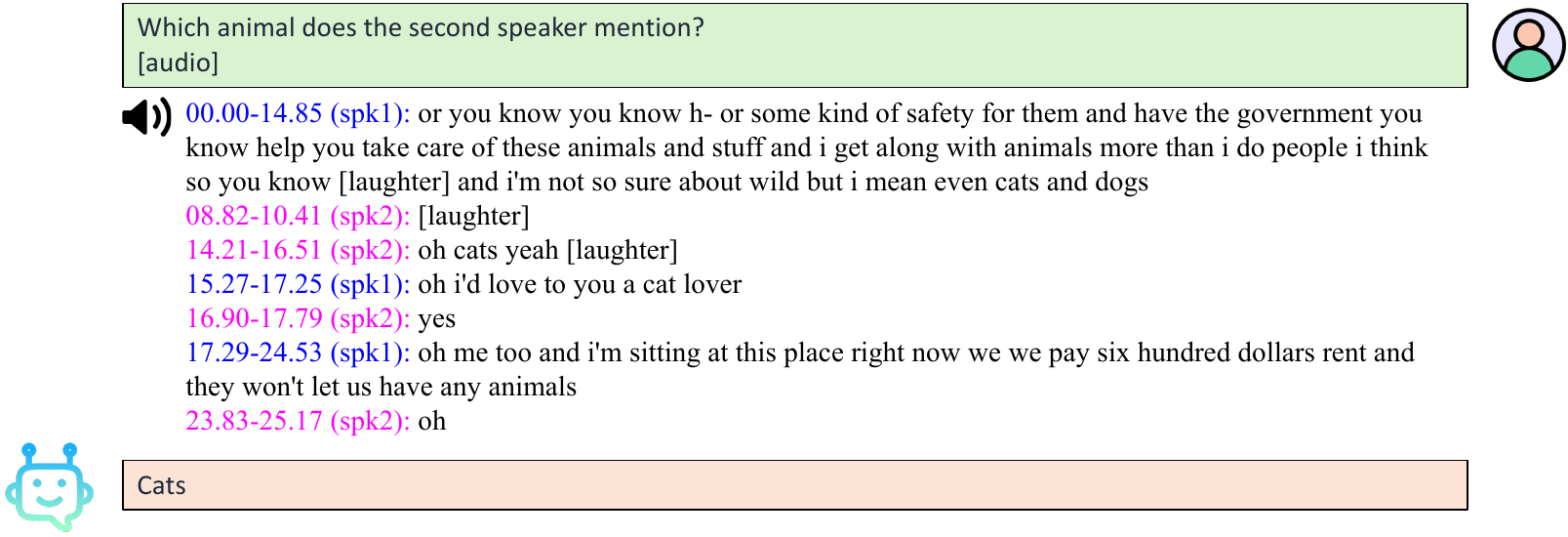}
          \vskip -0.1in
          \caption{Example 2}
          \label{subfig:demo-multispk2}
     \end{subfigure}
     \vskip -0.05in
     \caption{
     Our \method~can understand some multi-speaker dialog, despite being trained on single-speaker data only.
     }
     \label{fig:demo-multispk}
\end{figure*}

\section{Additional Demonstrations}

\subsection{Standard Speech Tasks}
\label{app:demo-standard}

\Cref{fig:demo-standardtasks} presents examples of ASR, AST, and SQA tasks. Our \method~performs well on those standard tasks.

\subsection{Generalization to Unseen Conditions}
\label{app:demo-unseen}

\Cref{fig:demo-contextualbiasing}, \Cref{fig:demo-math}, \Cref{fig:demo-coding}, and \Cref{fig:demo-multispk} are examples of contextual biasing ASR, math, coding, and SQA with multiple speakers.

\noindent\textbf{Contextual biasing.} \Cref{fig:demo-contextualbiasing} shows that the model can utilize additional contextual information when performing ASR. This can enhance ASR performance in specific domains without updating model parameters.

\noindent\textbf{Math w/ mixed-modal input.} \Cref{fig:demo-math} is an example of solving a math question using the information from both speech and text. The answer is correct and well formated, demonstrating that our \method~well preserves LLM's original capabilities.

\noindent\textbf{Coding.} \Cref{fig:demo-coding} is a coding example, where the model generates a correct response to the spoken instruction. Again, this shows that our \method~maintains the original LLM's capabilities in different domains.

\noindent\textbf{SQA w/ multi-speaker input.} 
The training data has only one speaker, but our model can also understand some multi-speaker conversations. In \Cref{fig:demo-multispk}, the audio contains two speakers with overlap. Our model can answer some questions related to different speakers correctly.

\end{document}